\documentclass{article}

\PassOptionsToPackage{numbers}{natbib}

\usepackage[final]{neurips_2022}

\usepackage[utf8]{inputenc} %
\usepackage[T1]{fontenc}    %
\usepackage{hyperref}       %
\usepackage{url}            %
\usepackage{booktabs}       %
\usepackage{amsfonts}       %
\usepackage{nicefrac}       %
\usepackage{microtype}      %
\usepackage{xcolor}         %

\usepackage{bm}
\usepackage{algorithm}
\usepackage{algpseudocode}
\usepackage{mathtools} %
\usepackage{dsfont} %
\usepackage{bbm}
\usepackage{booktabs} %
\usepackage{diagbox} 
\usepackage{nccmath} %
\usepackage{arydshln} %
\usepackage{tablefootnote}
\usepackage{amssymb}
\usepackage{enumitem} %
\usepackage{subfigure}
\usepackage{amsmath}
\usepackage{physics}

\usepackage{xcolor}
\usepackage{caption}
\usepackage{wrapfig}
\usepackage{multirow}

\usepackage{soul}

\usepackage{pifont} %

\usepackage{tablefootnote}

\usepackage{nccmath} %

\usepackage{graphicx} %

\newcommand{\oursfull}{Inverse Probability Weighting Distillation}
\newcommand{\ours}{IPWD}
\newcommand{\cmark}{\ding{51}}%
\newcommand{\domaint}{\textit{\text{machine}}}
\newcommand{\domains}{\textit{\text{human}}}
\newcommand{\Loss}{\mathcal{L}}
\newcommand{\loss}{\ell}
\newcommand{\ygt}{y}
\newcommand{\yt}{y^{t}}
\newcommand{\ys}{y^{s}}

\newcommand{\zs}{z^{s}}
\newcommand{\ykd}{y^{kd}}
\newcommand{\ycls}{y^{cls}}
\newcommand{\XEKD}{H(\ykd, y)}
\newcommand{\XECLS}{H(\ycls, y)}

\newcommand{\ie}{\textit{i.e.}}
\newcommand{\eg}{\textit{e.g.}}

\newcommand{\etal}{\textit{et al.}}

\title{Respecting Transfer Gap in Knowledge Distillation}

\author{%
  Yulei Niu\thanks{Work done when Yulei was at Nanyang Technological University.}~~$^1$
  \quad Long Chen$^1$ \quad Chang Zhou$^2$ \quad Hanwang Zhang$^3$
  \\
  $^1$Columbia University\quad $^2$Damo Academy, Alibaba Group\quad $^3$Nanyang Technological University\\
  \texttt{\{yn.yuleiniu,zjuchenlong\}@gmail.com}\quad \texttt{zhouchang.zc@alibaba-inc.com} \\
  \texttt{hanwangzhang@ntu.edu.sg}\\
}

\begin{document}

\maketitle

\begin{abstract}
    Knowledge distillation (KD) is essentially a process of transferring a teacher model's behavior, \eg, network response, to a student model. 
    The network response serves as additional supervision to formulate the machine domain ($\domaint$ for short), which uses the data collected from the human domain ($\domains$ for short) as a transfer set.
    Traditional KD methods hold an underlying assumption that
    the data collected in both human domain and machine domain are both independent and identically distributed (IID). We point out that this na\"ive assumption is unrealistic and there is indeed a transfer gap between the two domains. Although the gap offers the student model external knowledge from the machine domain, the imbalanced teacher knowledge would make us
    incorrectly estimate how much to transfer from teacher to student per sample on the non-IID transfer set. 
    To tackle this challenge, we propose \oursfull~(\ours) that estimates the propensity score of a training sample belonging to the machine domain, and assigns its inverse amount to compensate for under-represented samples. Experiments on CIFAR-100 and ImageNet demonstrate the effectiveness of \ours~for both two-stage distillation and one-stage self-distillation.
\end{abstract}
\section{Introduction}\label{sec:intro}

Knowledge distillation (KD)~\cite{hinton2015distilling} transfers knowledge from a teacher model, \eg, a big, cumbersome, and energy-inefficient network, to a student model, \eg, a small, light, and energy-efficient network, to improve the performance of the student model. 
A common intuition is that a teacher with better performance will teach a stronger student. However, recent studies find that the teacher's accuracy is not a good indicator of the resultant student performance~\cite{cho2019efficacy}. For example, a poorly-trained teacher with early stopping can still teach a better student~\cite{cho2019efficacy,dong2019distillation,yuan2020revisiting}; or, a teacher with a smaller model size than the student is also a good teacher~\cite{yuan2020revisiting}; or, a teacher with the same architecture as the student helps to improve the student---self-distillation~\cite{furlanello2018born,zhang2019your,zhang2021self,ji2021refine}. 

\begin{figure}
\centering
\includegraphics[width=0.8\textwidth]{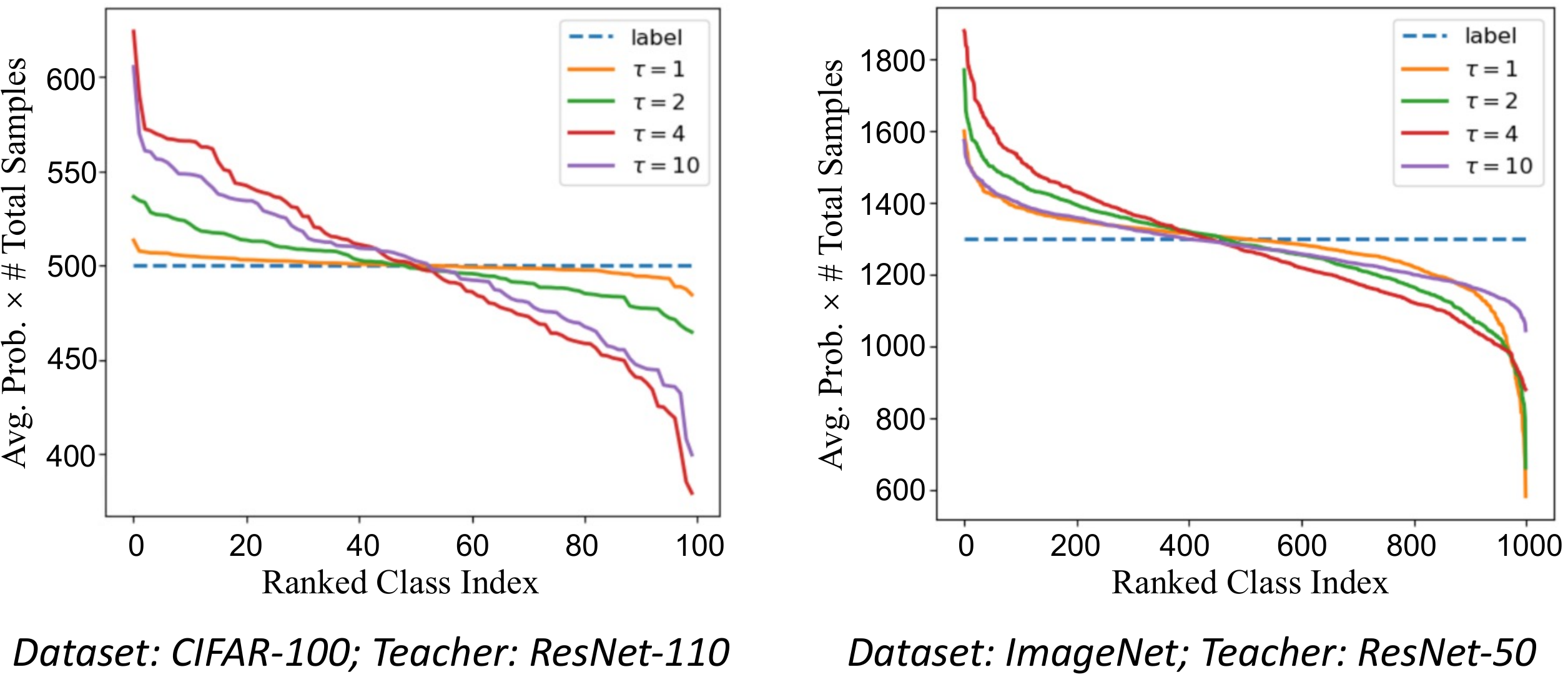}
\caption{Illustration of the distribution discrepancies among ground-truth annotations and teacher predictions.
Although 
the teacher model is trained on the balanced data (blue dashed),
its prediction distributions are imbalanced according to various temperatures.}
\label{fig:teaser}
\vspace{-4mm}
\end{figure}

Should we view KD in a perspective of domain transfer~\cite{duan2012domain, tan2017distant}, we would better understand the above counter-intuitive findings. From Figure~\ref{fig:teaser}, we can see that teacher predictions and ground-truth labels indeed behave differently. Although the teacher is trained on the balanced dataset, its predicted probability distribution over the dataset is imbalanced. 
Even on the same training set with the same model parameter, teachers with different temperature $\tau$ yield different ``soft label'' distributions from the ground-truth ones.  
This implies that human and teacher knowledge are from different domains, and there is a transfer gap that drives the ``dark knowledge''~\cite{hinton2015distilling} transferring from teacher to student---regardless of ``strong'' or ``weak'' teachers, it is a valid transfer as long as there is a gap.
{However, the transfer gap affects the distillation performance on the under-represented classes, \ie, classes on the tail of teacher predictions, which is overlooked in recent studies. Take CIFAR-100 as an example. We rank and divide the 100 classes into 4 groups according to the ranks of predicted probability. As shown in Table~\ref{tab:cifar100-teaser}, compared to vanilla training, KD achieves better performance in all the subgroups. However, the increase in the top 25 classes is much higher than that in the last 25 classes, \ie, averagely 5.14\% vs. 0.85\%.}
We ask: what causes the gap from the first place; or more specifically, why does the teacher's non-uniform distributed predictions implies the gap? We answer in an \emph{invariance vs. equivariance} learning point of view~\cite{arjovsky2019invariant, wang2021self}:

\begin{wraptable}{r}{9.0cm}
    \vspace{-4mm}
    \caption{{Improvement of KD over vanilla student for different classes. The metric is macro-average recall.}}
    \label{tab:cifar100-teaser}
    \centering
    \scalebox{0.75}{
    \begin{tabular}{l cccc}
        \toprule
         Arch. style & \textbf{Top 1-25} & \textbf{Top 26-50} & \textbf{Top 51-75} & \textbf{Top 76-100} \\
        \midrule
        \small ResNet50 -> MobileNetV2     & +4.96 & +5.92 & +1.76 & +1.20\\
        \small resnet32x4 -> ShuffleNetV1  & +5.80 & +2.68 & +2.52 & +0.84\\
        \small resnet32x4 -> ShuffleNetV2  & +4.72 & +1.92 & +2.24 & +0.76\\
        \small WRN-40-2 -> ShuffleNetV1    & +5.08 & +7.20 & +4.48 & +0.60\\
        \bottomrule
    \end{tabular}
    }
    \vspace{-4mm}
\end{wraptable} 

\textbf{Human domain: context invariance}. The discriminative generalization is the ability to learn both context-invariant and class-equivariant information from the diverse training samples per class. The human domain only provides context-invariant class-specific information, \ie, hard targets. We normally collect a balanced dataset to formulate human domain.

\textbf{Machine domain: context equivariance}. Teacher models often use a temperature variable to preserve the context. 
The temperature allows the teacher to represent a sample not only by its context-invariant class-specific information, but also its context-equivariant information. For example, a \texttt{dog} image with soft label 0.8$\cdot$\texttt{dog}~+~0.2$\cdot$\texttt{wolf} may imply that the dog has wolf-like contextual attributes such as ``fluffy coat'' and ``upright ears''. Although the context-invariance (\ie, class) is balanced in the training data, the context-equivariance (\ie, context) is imbalanced because the context balance is not considered in class-specific data collection~\cite{torralba2011unbiased}. 
To construct the transfer set for the machine domain, the teacher model annotates each sample after seeing others, \ie, being pre-trained on the whole set.
Interestingly, 
the diverse context results in a long-tailed imbalanced distribution, which is exactly reflected in Figure~\ref{fig:teaser}. 
In other words, the teacher's knowledge is imbalanced even though the teacher is trained on a class-balanced dataset.

Now we are ready to point out how the transfer gap is not properly addressed in conventional KD methods. 
Conventional KD calculates the Cross-Entropy (CE) loss between the ground-truth label and student's prediction, and the Kullback–Leibler (KL) divergence~\cite{kullback1951information} loss between the teacher's and student's predictions, where a constant weight is assigned for the two losses.
This is essentially based on the underlying assumption that 
the data in both the human and machine domains are IID.
Based on the analysis of context equivariance, we argue that the assumption is unrealistic, \ie, \textit{the teacher's knowledge is imbalanced}. Therefore, a constant sample weight for the KL loss would be a bottleneck.
In this paper, we propose a simple yet effective method, \oursfull~(\ours),
which compensates for the training samples that are under-weighted in the machine domain. For each training sample $x$, we first estimate its machine-domain propensity score $P(x|\domaint)$ by
comparing class-aware and context-aware predictions. 
A sample with a low propensity score would have a high confidence from class-aware predictions and a low confidence from context-aware predictions.
Then, \ours~assigns the inverse probability 
$1/P(x|\domaint)$ 
as the sample weight for the KL loss to highlight the under-represented samples. 
In this way, IPWD generates a pseudo-population~\cite{little2019statistical,imbens2015causal} to deal with the imbalanced knowledge.

We evaluate our proposed \ours~on two typical knowledge distillation settings: two-stage teacher-student distillation and one-stage self-distillation. Experiments conducted on  CIFAR-100~\cite{krizhevsky2009learning} and ImageNet~\cite{deng2009imagenet} demonstrate the effectiveness and generality of our \ours. 

Our contributions are three-fold:
\vspace{-2mm}
\begin{itemize}[leftmargin=*]
\item We formulate KD as a domain transfer problem and argue that the na\"ive IID assumption on machine domain neglects the imbalanced knowledge due to transfer gap.
\item We propose \oursfull~(\ours)~which compensate for the samples that are overlooked in the machine domain to tackle the imbalanced knowledge in transfer gap.
\item Experiments on CIFAR-100 and ImageNet for both two-stage distillation and one-stage self-distillation show that the proper handling of the transfer gap is a promising direction in KD. 
\end{itemize}
\section{Related Work}

\noindent \textbf{Knowledge distillation} (KD) was first introduced to transfer the knowledge from an effective but cumbersome model to a smaller and efficient model~\cite{hinton2015distilling}. The knowledge can be formulated in either output space~\cite{hinton2015distilling,jin2019knowledge,li2020local,yun2020regularizing,yuan2020revisiting,mirzadeh2020improved,son2021densely,zhou2021rethinking,kim2021self} or representation space~\cite{romero2014fitnets,huang2017like,zagoruyko2016paying,kim2018paraphrasing,park2019relational,heo2019comprehensive,tian2019contrastive,chen2021distilling,ji2021refine}. KD has attracted a wide interest in theory, methodology, and applications~\cite{gou2021knowledge}. For applications, KD has shown its great potential
in various areas, including but not limited to classification~\cite{li2017learning,peng2019few,luo2018graph,hu2021distilling}, detection~\cite{li2017mimicking,shmelkov2017incremental,wang2019distilling}, segmentation~\cite{he2019knowledge,mullapudi2019online,liu2019structured} for visual recognition tasks, and visual question answering~\cite{mun2018learning,aditya2019spatial,niu2021introspective}, video captioning~\cite{pan2020spatio,zhang2020object}, and text-to-image synthesis~\cite{tan2020kt} for vision-language tasks. Recent studies further discussed how and why KD works. Specifically, M\"uller \etal~\cite{muller2019does} and Shen \etal~\cite{shen2020label} empirically analyzed the effect of label smoothing on KD. Cho \etal~\cite{cho2019efficacy}, Dong \etal~\cite{dong2019distillation}, and Yuan \etal~\cite{yuan2020revisiting} pointed out that early stopping is a good regularization for a better teacher. Yuan \etal~\cite{yuan2020revisiting} further found that a poorly trained teacher, even a model smaller than the student, can improve the performance of the student. Besides, Memon \etal~\cite{menon2021statistical} and Zhou \etal~\cite{zhou2021rethinking} proposed a bias-variance trade-off perspective for KD. In this paper, we point out that existing KD methods hold an underlying assumption that the IID training samples are also IID in the machine domain, which overlooks the transfer gap.

\noindent \textbf{Self-distillation} is a special case of KD, which uses the student network itself as the teacher instead of the cumbersome model, \ie, the teacher and student models have the same architecture~\cite{furlanello2018born,zhang2019your,zhang2021self,ji2021refine}. This process can be executed in iterations and produce a stronger ensemble model~\cite{furlanello2018born}. Similar to KD, traditional self-distillation follows a two-stage process: first pre-training a student model as the teacher, and then distilling the knowledge from the pre-trained model to a new student model. In order to perform the teacher-student optimization in one generation, recent studies~\cite{yang2019snapshot,kim2021self} proposed one-stage self-distillation that adopts student models at earlier epochs as teacher models. These one-stage self-distillation methods outperform vanilla students by large margins. In this paper, we also evaluate the effectiveness of our \ours~as a plug-in in one-stage self-distillation.

\noindent \textbf{Inverse Probability Weighting} (IPW)~\cite{rosenbaum1983central,little2019statistical,imbens2015causal,austin2011introduction}, also known as inverse probability of treatment weighting or inverse propensity weighting, was proposed to correct the selection bias when the observations are non-IID. IPW uses the inverse of the probability (\ie, propensity score) that the individual would be assigned to the treatment group to reweight the samples. Propensity-weighting techniques have been widely applied and studied in many areas~\cite{schnabel2016recommendations}, such as causal inference~\cite{imbens2015causal}, complete-case analysis~\cite{little2019statistical}, machine learning~\cite{cortes2008sample,bickel2009discriminative,sugiyama2012machine}, and recommendation systems~\cite{schnabel2016recommendations,wang2018deconfounded,ai2018unbiased}.
In this paper, we view the distillation process as a domain transfer problem and adopt IPW to dynamically assign the weight to each training sample for the distillation loss. 

\section{Analysis}\label{sec:background}

\subsection{Knowledge Distillation (KD)}

We view knowledge distillation from a perspective of domain transfer, and take the image classification task as the case study. Suppose that the training data $\mathcal{D}\!=\{\mathcal{X},\mathcal{Y}\}\!=\!\{(x,\ygt)\}$ contains $x$ as the input (\eg, image) and $\ygt\!\in\!\mathbb{R}^C$ as its ground-truth annotation (\eg, one-hot label), where $C$ denotes the number of classes. A standard solution to train the classifier $\theta$ uses the cross-entropy loss as the objective:
\begin{align}
    \Loss_{cls}(\domains;\theta)&=\mathbb{E}_{(x,y)\sim P_{\domains}}[\loss_{cls}(x,\ygt;\theta)]\approx \frac{1}{|\mathcal{D}|}\sum_{(x,\ygt)\in\mathcal{D}}\loss_{cls}(x,\ygt;\theta)
    \triangleq\Loss_{cls}(\mathcal{D};\theta),\label{eq:cls1}
\end{align}
where $\loss_{cls}(x,\ygt)\!=\!H(\ys,\ygt)$ is the classification loss for sample $x$, $H(p,q)=\sum^{C}_{i=1}-q_i\log p_i$ denotes the cross entropy between $p$ and $q$, $\ys\!=\!f(x;\theta)$ denotes the model's output probability given $x$, \ie, $y^s_k\!=\!\frac{\exp(z^s_k)}{\sum^{C}_{i=1}\exp(z^s_i)}$, where $\zs$ is the output logits of the model. The hard targets provide context-invariant class-specific information from the human domain. 
An assumption held behind Eq.~\eqref{eq:cls1} is that
the samples are independent and identically distributed (IID) in the training and test set. 

KD adopts a teacher model $\theta^t$ to generate soft targets as extra supervisions, \ie, context-equivariant information. To formulate the machine domain, traditional KD methods commonly use the training set $\mathcal{D}$ to construct the transfer set
$\mathcal{D}^{t}$ using the same copy of $\mathcal{X}$, \ie, $\mathcal{D}^{t}\!=\!\{(x,\yt)\}$ where $\yt\!=\!f(x;\theta^t)$ and $x\in\mathcal{X}$. Traditional KD approaches use the KL divergence~\cite{kullback1951information} loss for knowledge transfer:
\begin{align}
    \Loss_{dist}(\domaint;\theta)
    =\mathbb{E}_{(x,y)\sim P_{\domaint}}[\loss_{dist}(x,\ygt;\theta)]
    \approx\frac{1}{|\mathcal{D}^{t}|}\sum_{(x,\yt)\in\mathcal{D}^{t}}\loss_{dist}(x,\yt;\theta)
    \triangleq\Loss_{dist}(\mathcal{D}^{t};\theta),\label{eq:kd1}
\end{align}
where $\loss_{dist}(x,\yt;\theta)\!=\tau^2\cdot\![H(\ys_\tau,\yt_\tau)-H(\yt_\tau,\yt_\tau)]$ 
denotes the distillation loss for sample $x$. Normally, the outputs of the student and teacher are softened using a temperature $\tau$, \ie, $y^s_{\tau,k}\!=\!\frac{\exp(z^s_k/\tau)}{\sum^{C}_{i=1}\exp(z^s_i/\tau)}$ and $y^t_{\tau,k}\!=\!\frac{\exp(z^t_k/\tau)}{\sum^{C}_{i=1}\exp(z^t_i/\tau)}$. The overall objective combines $\Loss_{cls}$ and $\Loss_{dist}$ as:
\begin{equation}    \Loss_{kd}=\alpha\cdot\Loss_{cls}+\beta\cdot\Loss_{dist},
\end{equation}
where $\alpha$ and $\beta$ are the hyper-parameters. The underlying assumption of traditional KD behind Eq.~\eqref{eq:kd1} is that
the transfer set $\mathcal{D}^{t}$ is an unbiased approximation of the machine domain.
However, the observed long-tailed and temperature-sensitive distributions of teacher's predictions in Figure~\ref{fig:teaser} rationally challenge this assumption. 
As a result, samples with lower $P(x|\domaint)$ are under-represented during the distillation process, which affects the unbiasedness of knowledge transfer.
This analysis indicates that Eq.~\eqref{eq:kd1} is not optimal to utilize the teacher's imbalanced knowledge.

\subsection{{Transfer Gap in KD}}\label{sec:transfergap}

\begin{wrapfigure}{r}{0.35\textwidth}
\centering
\vspace{-12mm} 
\includegraphics[width=0.22\textwidth]{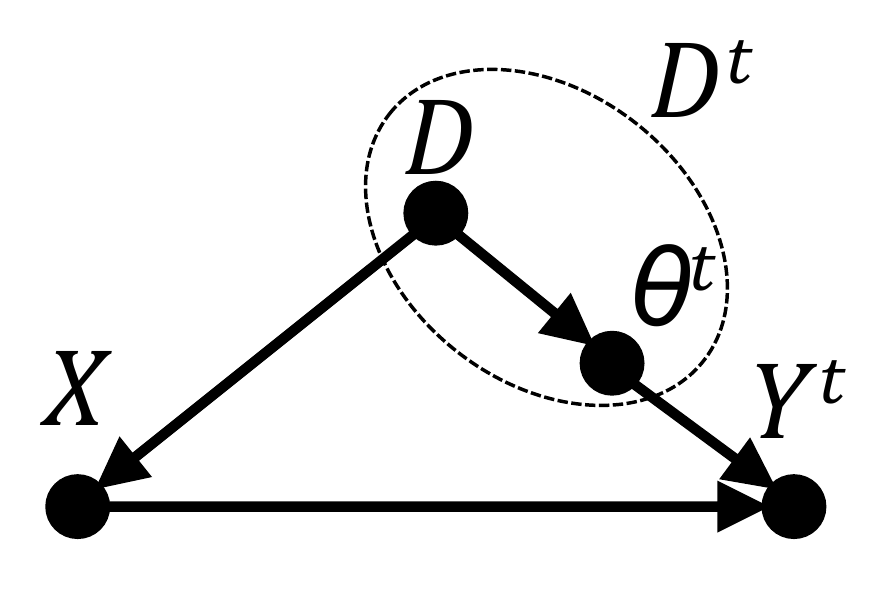}
\caption{Causal graph for KD.}
\label{fig:causal-graph}
\vspace{-3mm}
\end{wrapfigure}

We interpret the transfer gap and its confounding effect from the perspective of causal inference. Figure~\ref{fig:causal-graph} illustrates the causal relations between the image $X$, training data $\mathcal{D}=\{(x,y)\}$, teacher's parameters $\theta^t$ and teacher's output $Y^t$ in KD. Overall, $\mathcal{D}$ and $\theta^t$ jointly act as the confounder of $X$ and $Y^t$ in the transfer set. First, the training set $\mathcal{D}$ and transfer set of teacher model $\mathcal{D}^t=\{(x,y^t)\}$ share the same image set, and $X=x$ is sampled from the image set of $\mathcal{D}$, i.e., $\mathcal{D}$ serves the cause of $X$. Second, the teacher $\theta^t$ is trained on $\mathcal{D}$, and $y^t$ is calculated based on $\theta^t$ and $x$, i.e., $y^t=f(x;\theta^t)$. Therefore, $X$ and $\theta^t$ are the cause of $Y^t$. Note that the transfer set is constructed based on the images on $\mathcal{D}$ and teacher model $\theta^t$. Therefore, we regard the transfer set $\mathcal{D}^t$, the joint of $\mathcal{D}$ and $\theta^t$, as the confounder of $X$ and $Y^t$.

Although $\mathcal{D}$ is balanced when considering the context-invariant class-specific information, the context information (e.g., attributes) is overlooked, which makes the $\mathcal{D}$ imbalanced on context. As shown in Figure~\ref{fig:teaser}, such imbalanced context leads to an imbalanced transfer set $\mathcal{D}^t$ and further affects the distillation performance of teacher's knowledge.

To overcome such confounding effect, a commonly used technique is intervention via $P(y^t|do(x))$ instead of $P(y^t|x)$, which is formulated as $P(y^t|do(x))=\sum_{\mathcal{D}^t} P(y|x,\mathcal{D}^t)P(\mathcal{D}^t)=\sum_{\mathcal{D}^t} \frac{P(x,y^t,\mathcal{D}^t)}{P(x|\mathcal{D}^t)}$. This transformation suggests that we can use the inverse of propensity score,  $1/P(x|\mathcal{D}^t)$, as sample weight to implement the intervention and overcome the confounding effect. Thanks to the causality-based theory~\cite{rosenbaum1983central,austin2011introduction}, we can use the Inverse Probability Weighting (IPW) technique to overcome the confounding effect brought by the transfer gap.
\section{Method}\label{sec:method}

We propose a simple yet effective method, \oursfull~(\ours), to respect the transfer gap and imbalance knowledge in KD. In this section, we first introduce the overall framework of \ours, then present the implementation details.

\subsection{Inverse Probability Weighting for KD}

As analyzed in Section~\ref{sec:background}, the IID training samples in the human domain are no longer IID in the machine domain.
Simply assuming the training set as the perfect transfer set may lead to the selection bias: samples that match ``head'' knowledge are over-represented and easy to be observed, while samples that match ``tail'' knowledge are under-represented and hard to be observed. This would suppress the transfer of ``tail'' knowledge. The analysis from the perspective of causal inference in Section~\ref{sec:transfergap} suggests that we can use Inverse Probability Weighting (IPW) for debiased distillation. In short, IPW generates a pseudo-population where under-represented samples are assigned with large weights and over-represented samples are assigned with small weights. The weight for sample $x$ is determined as the inverse of its probability, also known as propensity score, to the domain $d\in\{\domains,\domaint\}$, \ie, $w_{x|d}\!=\!1/p(x|d)$. We adopt IPW to KD and obtain the following objective for sample $x$:
\begin{equation}
    \loss(x;\theta)=\sum_{d}w_{x|d}\cdot\loss_d(x,y^d;\theta)=\frac{1}{P(x|\domains)}\loss_{cls}(x,\ygt;\theta)+\frac{1}{P(x|\domaint)}\loss_{dist}(x,\yt;\theta).
\end{equation}

\begin{figure}
\centering
\includegraphics[width=0.9\textwidth]{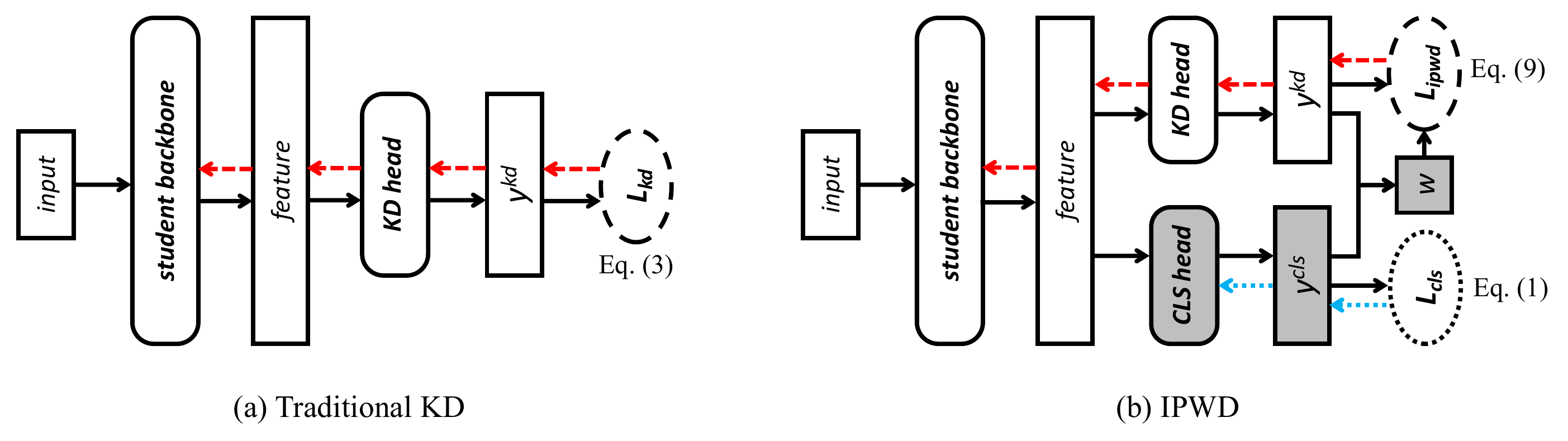}
\caption{The comparison of training pipelines between traditional KD and our \ours. 
The module and outputs in grey are not used in the delivered student model.}
\label{fig:framework}
\end{figure}

\subsection{Implementation}\label{sec:method-imp}
Since the training and test data are normally IID in the human domain, we safely and rationally use the empirical risk.
Therefore, we assign a constant weight to each sample when calculating the classification loss. For sure, the training and test samples can be both non-IID, \eg, long-tailed recognition tasks, which is out of the scope of this paper. 

As analyzed in Section~\ref{sec:intro}, the assumption held by traditional KD, \ie, 
both $P(X|\domains)$ and $P(X|\domaint)$ are IID,
is unrealistic in practice. Therefore, we should consider the propensity score $P(x|\domaint)$ as a sample-specific value for the distillation loss to improve the generalization. Traditional IPW estimates the propensity score using logistic regression, \ie, $\hat{P}(x|\domaint)=1/(1+\exp(-z_x))$, where $z_x$ is the logit for $x$. Since the ground-truth annotation of $P(x|\domaint)$ is not available, it is not practicable to directly train the regression model in a fully-supervised manner. Therefore, we estimate the propensity score in an unsupervised way.

Recall that the samples with high propensity are over-represented in the transfer set. As a result, the student model would learn less from the under-represented group via distillation. Therefore, we use a classification-trained (CLS-trained) classifier for the human domain as reference, and assume that a KD-trained classifier for the machine domain is more confident for the over-represented group than the CLS-trained classifier.
We compare the outputs of two classifiers to identify whether a sample is under-represented in the machine domain. 
Suppose that the KD-trained output is $\ykd$ and the CLS-trained output is $\ycls$. The assumption implies that the logit $z_x$ is negatively correlated with $H(\ykd,y)$ and positively correlated with $H(\ycls,y)$, where $H(\cdot,\cdot)$ is the cross-entropy defined in Section~\ref{sec:background}. Considering the range of logit, we estimate $z_x$ as $z_x=\log\frac{\XECLS}{\XEKD}$.

\begin{figure*}
\centering
\includegraphics[width=0.95\textwidth]{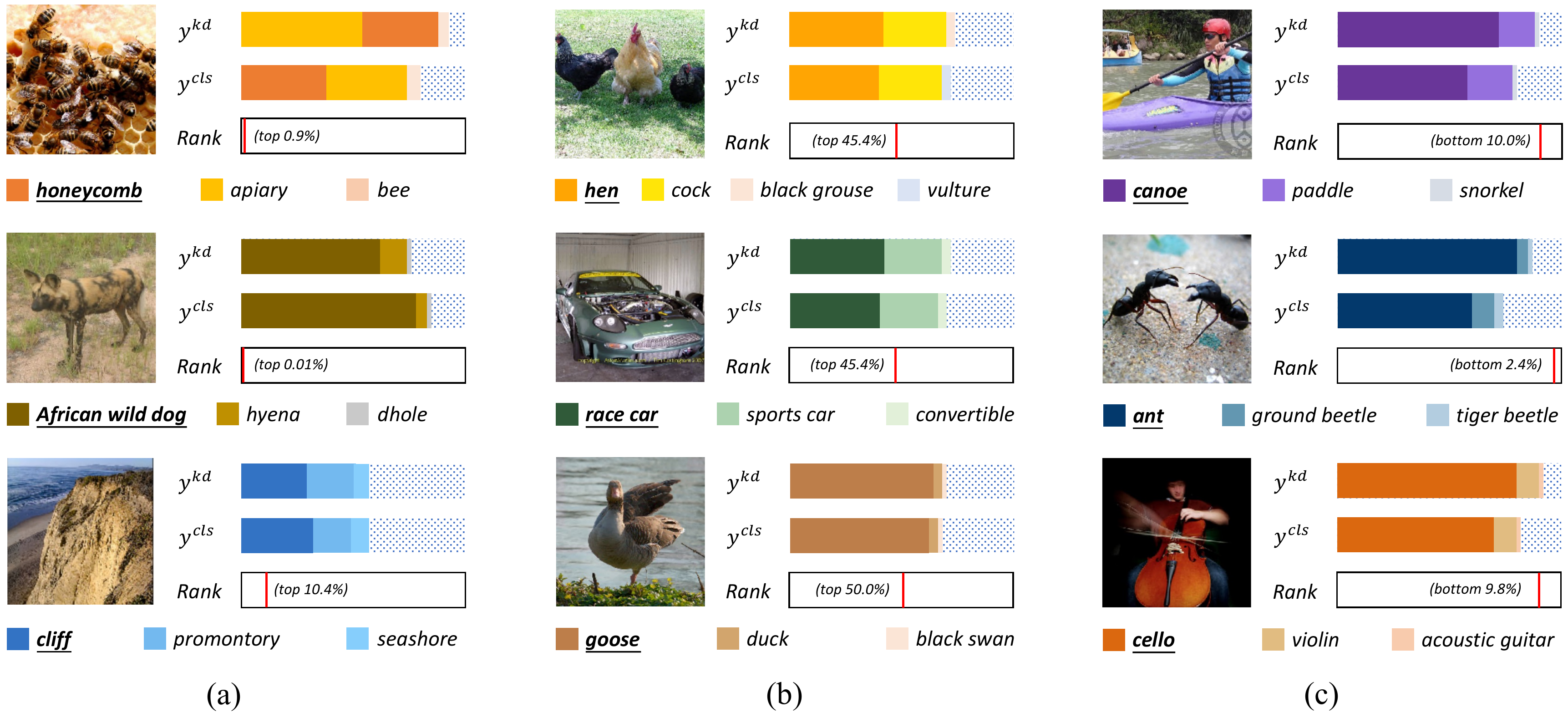}
\vspace{-2mm}
\caption{Illustration of the sample weights. \textbf{\underline{Bold underlined class}} denotes the ground-truth label. The areas in $y^{kd}$ and $y^{cls}$ represent the proportion of the predicted probability. A high rank indicates a large sample weight, while a low rank indicates a small sample weight.}
\label{fig:examples}
\end{figure*}

Figure~\ref{fig:framework} illustrates the comparison between traditional KD and our \ours. We take the KD-trained student's output $\ys$ as $\ykd$, and train an extra classifier head (\ie, ``CLS head'' in Figure~\ref{fig:framework}(b)) to calculate $\ycls$, which is optimized using the cross-entropy loss $\Loss_{cls}$ with the ground-truth labels. As shown in Table~\ref{tab:cifar100-ab-component}, we empirically found that directly using $\ys$ and $\ycls$ would lead to a high variance. For example, a wrongly classified sample may have a extremely large loss and heavily suppress the distillation of other samples. Therefore, we normalize the logits by dividing them by the standard deviation $\sigma^s$ and $\sigma^{cls}$, \ie, $\widetilde{y}^s\!=\!\frac{\exp(z^s_k/\sigma^s)}{\sum^{C}_{i=1}\exp(z^s_i/\sigma^s)}$ and $\widetilde{y}^{cls}\!=\!\frac{\exp(z^{cls}_k/\sigma^{cls})}{\sum^{C}_{i=1}\exp(z^{cls}_i/\sigma^{cls})}$. In this way, the outputs are at the same scale with a standard deviation equal to 1, which helps to reduce the variance. 
We finally take $\widetilde{y}^s$ as $\ykd$ and $\widetilde{y}^{cls}$ as $\ycls$.
Combining $z_x$ into the propensity score, we have:
\begin{equation}
    \hat{P}(x|\domaint)=\frac{\XECLS}{\XECLS+\XEKD},
\end{equation}
and the estimated weight $\hat{w}_{x}$ for sample $x$ is $\hat{w}_{x}=\frac{1}{\hat{P}(x|\domaint)}=1+\frac{\XEKD}{\XECLS}$.
Figure~\ref{fig:examples} illustrates some examples of training samples and their assigned weights. Under-represented samples, for which the KD-trained classifier is less confident than the CLS-trained classifier, are assigned with a large weight (Figure ~\ref{fig:examples}(a)). Over-represented samples, for which the KD-trained classifier is more confident than the CLS-trained classifier, are assigned with a small weight (Figure ~\ref{fig:examples}(c)). Samples for which the two classifiers behave similarly are assigned with a balanced weight (Figure ~\ref{fig:examples}(b)). The weighted distillation loss is formulated as:
\begin{equation}
    \Loss_{ipw\text{-}dist}=\frac{1}{|\mathcal{D}^t|}\sum_{(x,y^t)\in\mathcal{D}^t}\hat{w}_{x}\cdot\loss_{dist}(x,y^t;\theta),
\end{equation}

Our final \ours~objective is formulated as:
\begin{equation}
    \Loss_{ipwd}=\Loss_{cls}+\alpha\Loss_{ipw\text{-}dist}\label{eq:final}
\end{equation}
where $\alpha$ is a trade-off hyper-parameter between classification and distillation.

\noindent \textbf{Limitations and negative societal impacts.}. As introduced in Section~\ref{sec:method-imp}, we estimate the propensity score by comparing the heads of the student model. Therefore, the estimation relies on the quality of the student model. A poor student may not correctly estimate the propensity, which may further suppress the effectiveness of \ours. 
Also, we hold an assumption that the training and test samples are IID in the human domain, which may not be valid for long-tailed tasks. To the best of our knowledge, as our work is purely an algorithm for knowledge distillation, we haven't found any negative societal impact.
\section{Experiments}\label{sec:experiments}

We take the image classification task as a case study to evaluate the effectiveness and generalizability of our \ours. Following previous works~\cite{tian2019contrastive,zhou2021rethinking,kim2021self}, we conduct experiments with two settings, two-stage distillation and one-stage self-distillation.

\subsection{Datasets and Settings}

\textbf{Datasets}. 
We conducted experiments on CIFAR-100~\cite{krizhevsky2009learning} and ImageNet~\cite{deng2009imagenet}. 
CIFAR-100 contains 50K images in the training set and 10K images in the test set from 100 classes. ImageNet provides 1.2M images in the training set and 50K images in the validation set from 1K classes.

\textbf{Settings}. Two-stage distillation is the conventional setting that pre-trains a teacher model at the first stage and transfers the knowledge to a student model at the second stage. Commonly, the teacher is a larger model, and the student is a smaller model. For self-distillation, the teacher and student have the same architecture. One-stage self-distillation aims to complete the teacher-student optimization simultaneously~\cite{yang2019snapshot,kim2021self}, \ie, the pre-training and transfer processes are reduced to one.

\subsection{Two-stage Distillation}

\textbf{Baseline methods}. For two-stage distillation, following Tian \etal~\cite{tian2019contrastive} and Zhou \etal~\cite{zhou2021rethinking}, we considered the following methods as baselines: KD~\cite{hinton2015distilling}, FitNet~\cite{romero2014fitnets}, AT~\cite{zagoruyko2016paying}, SP~\cite{tung2019similarity}, CC~\cite{peng2019correlation}, VID~\cite{ahn2019variational}, RKD~\cite{park2019relational}, PKT~\cite{passalis2018learning}, FSP~\cite{yim2017gift}, AB~\cite{heo2019knowledge}, FT~\cite{kim2018paraphrasing}, NST~\cite{huang2017like}, CRD~\cite{tian2019contrastive}, SSKD~\cite{xu2020knowledge}, and WSLD~\cite{zhou2021rethinking}. In particular, WSLD~\cite{zhou2021rethinking} is the most related work to us, which proposed a bias-variance trade-off perspective for KD and also assigns different weights to each training sample. Similarly, the weight is positive related to the cross-entropy loss of student's output. The main differences between our \ours~and WSLD are as follows. First, our formulation of samples weights is theoretically guaranteed by the causal theory behind Inverse Probability Weighting (IPW)~\cite{rosenbaum1983central,little2019statistical,imbens2015causal,austin2011introduction}. Second, WSLD estimates the sample weight using both the student model and the teacher model. As a comparison, we use the student model with two different classifier heads to guarantee that the capacities of the compared models are close.

\begin{table*}
    \caption{Top-1 accuracies (\%) on CIFAR-100 for two-stage distillation. $^*$ denotes our reproduced results using the same teacher model.}
    \label{tab:cifar100}
    \centering
    \scalebox{0.93}{
    \begin{tabular}{l cccc cccc}
    \toprule
    & \multicolumn{4}{c}{Same architecture style} & \multicolumn{4}{c}{Different architecture style} \\ 
    \cmidrule(lr){2-5} \cmidrule(lr){6-9} 
    Teacher & {\scriptsize WRN-40-2} & {\scriptsize resnet56} & {\scriptsize resnet110} & {\scriptsize resnet32x4} & {\scriptsize resnet32x4} & {\scriptsize WRN-40-2} & {\scriptsize ResNet50} & {\scriptsize ResNet50} \\
    Student & {\scriptsize WRN-40-1} & {\scriptsize resnet20} & {\scriptsize resnet32} & {\scriptsize resnet8x4} & {\scriptsize ShuffleNetV1} & {\scriptsize ShuffleNetV1} & {\scriptsize vgg8} & {\scriptsize MobileNetV2} \\
    \midrule
    Teacher & 75.61 & 72.34 & 74.31 & 79.42 & 79.42 & 75.61 & 79.34 & 79.34 \\
    Student & 71.98 & 69.06 & 71.14 & 72.50 & 70.50 & 70.50 & 70.36 & 64.60 \\ 
    \midrule
    FitNet~\cite{romero2014fitnets} & 72.24 & 69.06 & 71.06 & 73.50 & 73.59 & 73.73 & 70.69 & 63.16 \\
    AT~\cite{zagoruyko2016paying} & 72.77 & 69.21 & 72.31 & 73.44 & 71.73 & 73.32 & 71.84 & 58.58 \\
    SP~\cite{tung2019similarity} & 72.43 & 69.67 & 72.69 & 72.94 & 73.48 & 74.52 & 73.34 & 68.08 \\
    CC~\cite{peng2019correlation} & 72.21 & 69.63 & 71.48 & 72.97 & 71.14 & 71.38 & 70.25 & 65.43 \\
    VID~\cite{ahn2019variational} & 73.30 & 70.38 & 72.61 & 73.09 & 73.38 & 73.61 & 70.30 & 67.57 \\
    RKD~\cite{park2019relational} & 72.22 & 69.61 & 71.82 & 71.90 & 72.28 & 72.21 & 71.50 & 64.43 \\
    PKT~\cite{passalis2018learning} & 73.45 & 70.34 & 72.61 & 73.64 & 74.10 & 73.89 & 73.01 & 66.52 \\
    AB~\cite{heo2019knowledge} & 72.38 & 69.47 & 70.98 & 73.17 & 73.55 & 73.34 & 70.65 & 67.20 \\
    FT~\cite{kim2018paraphrasing} & 71.59 & 69.84 & 72.37 & 72.86 & 71.75 & 72.03 & 70.29 & 60.99 \\
    NST~\cite{huang2017like} & 72.24 & 69.60 & 71.96 & 73.30 & 74.12 & 74.89 & 71.28 & 64.96 \\
    KD~\cite{hinton2015distilling} & 73.54 & 70.66 & 73.08 & 73.33 & 74.07 & 74.83 & 73.81 & 67.35 \\
    CRD~\cite{tian2019contrastive} & 74.14 & 71.16 & 73.48 & 75.51 & 75.11 & 76.05 & 74.30 & 69.11 \\
    WSLD$^*$~\cite{zhou2021rethinking} & 73.74 & \bf 71.53 & 73.36 & 74.79 & 75.09 & 75.23 & 73.80 & 68.79 \\
    \midrule
    \bf \ours & \bf 74.64 & 71.32 & \bf 73.91 & \bf 76.03 & \bf 76.03 & \bf 76.44 & \bf 74.97 & \bf 70.25 \\
    \bottomrule 
    \end{tabular}
    }
\end{table*}

\begin{table*}
    \caption{Top-1 accuracies (\%) on CIFAR-100 test set as a plug-in on SSKD~\cite{xu2020knowledge}. We reproduced the results of SSKD using the same teacher model.}
    \label{tab:cifar100-sskd}
    \centering
    \scalebox{0.9}{
    \begin{tabular}{l cccc cccc}
    \toprule
    & \multicolumn{4}{c}{Same architecture style} & \multicolumn{4}{c}{Different architecture style} \\ 
    \cmidrule(lr){2-5} \cmidrule(lr){6-9} 
    Teacher & {\scriptsize WRN-40-2} & {\scriptsize WRN-40-2} & {\scriptsize resnet56} & {\scriptsize resnet32x4} & {\scriptsize ResNet50} & {\scriptsize resnet32x4} & {\scriptsize WRN-40-2} & {\scriptsize vgg13} \\
    Student & {\scriptsize WRN-16-2} & {\scriptsize WRN-40-1} & {\scriptsize resnet20} & {\scriptsize resnet8x4} & {\scriptsize MobileNetV2} & {\scriptsize ShuffleNetV1} & {\scriptsize ShuffleNetV1} & {\scriptsize MobileNetV2} \\
    \midrule
    Teacher & 76.46 & 76.46 & 73.44 & 79.63 & 79.10 & 79.63 & 76.46 & 75.38 \\
    Student & 73.64 & 72.24 & 69.63 & 72.51 & 65.79 & 70.77 & 70.77 & 65.79 \\ 
    \midrule
    SSKD$^*$~\cite{xu2020knowledge} & 75.74 & 75.59 & 70.61 & 75.80 & 72.22 & 77.71 & 78.49 & 77.32 \\
    \bf +~\ours & \bf 76.39 & \bf 76.09 & \bf 71.69 & \bf 76.74 & \bf 72.85 & \bf 78.30 & \bf 79.17 & \bf 77.95 \\
    \bottomrule 
    \end{tabular}
    }
\end{table*}

\textbf{Implementation}. For experiments on CIFAR-100, we followed CRD~\cite{tian2019contrastive} based on the open-sourced code. We set the trade-off hyper-parameter $\alpha\!=\!5$ in Eq.~\eqref{eq:final} and the temperature $\tau\!=\!10$. Other training details were the same as CRD~\cite{tian2019contrastive} and provided in the appendix.
For ImageNet, we followed Zhou \etal~\cite{zhou2021rethinking} to conduct experiments based on their open-sourced code. We used the same hyper parameters as WSLD~\cite{zhou2021rethinking}, \ie, $\alpha$ as 2.5 and $\tau$ as 2.

\textbf{Comparison with baseline methods}. Table~\ref{tab:cifar100} shows the results of student models on CIFAR-100 with different teacher-student architectures, which can be grouped into same architecture style and different architecture style. Note that the results of WSLD reported in~\cite{zhou2021rethinking} used a different pre-trained teacher model. Since some training techniques like early-stopping~\cite{cho2019efficacy,dong2019distillation,yuan2020revisiting} may improve the distillation performance, we reimplemented WSLD using the same teacher model for a fair comparison. 
Overall, our \ours~outperforms KD by large margins and outperforms other baseline methods on most of the architectures, which demonstrates the effectiveness of our \ours.
In particular, the improvement with the same architecture style is smaller than the different style. The reason is that the different architecture style reflects the bigger gap between the human domain and machine domain. Since our \ours~weights the training samples to address the non-IID problem, \ours~successfully outperforms KD and other state-of-the-art methods by large margins when the transfer gap is significant.

\begin{wraptable}{r}{8.0cm}
    \vspace{-4mm}
    \caption{Acc. (\%) on ImageNet for two-stage distillation.}
    \label{tab:imagenet-mobile}
    \centering
    \scalebox{0.98}{
    \begin{tabular}{l cc cc}
        \toprule
         & \multicolumn{2}{c}{Same arch. style} & \multicolumn{2}{c}{Diff. arch. style} \\
        \cmidrule(lr){2-3} \cmidrule(lr){4-5} 
        Teacher & \multicolumn{2}{c}{ResNet-34} & \multicolumn{2}{c}{ResNet-50} \\ 
        Student & \multicolumn{2}{c}{ResNet-18} & \multicolumn{2}{c}{MobileNet-v1} \\ 
        \midrule
        & \bf Top-1 & \bf Top-5 & \bf Top-1 & \bf Top-5 \\ 
        \midrule
        Teacher & 73.31 & 91.42 & 76.16 & 92.87\\
        Student & 69.75 & 89.07 & 68.87 & 88.76\\
        \midrule
        AT~\cite{zagoruyko2016paying} & 71.03 & 90.04 & 70.18 & 89.68\\
        NST~\cite{huang2017like} & 70.29 & 89.53 & --- & --- \\
        FT~\cite{kim2018paraphrasing} & --- & --- & 69.88 & 89.50 \\
        FSP~\cite{yim2017gift} & 70.58 & 89.61 & --- & --- \\
        AB~\cite{heo2019knowledge} & --- & --- & 68.89 & 88.71\\
        RKD~\cite{park2019relational} & 70.40 & 89.78 & 68.50 & 88.32\\
        KD~\cite{hinton2015distilling} & 70.67 & 90.04 & 70.49 & 89.92\\
        Overhaul~\cite{heo2019comprehensive} & 71.03 & 90.15 & 71.33 & 90.33\\
        CRD~\cite{tian2019contrastive} & 71.17 & 90.13 & 69.07 & 88.94 \\
        SSKD~\cite{xu2020knowledge} & 71.62 & 90.67 & --- & --- \\
        DGKD~\cite{son2021densely} & 71.73 & \bf 90.82 & --- & --- \\
        WSLD~\cite{zhou2021rethinking} & \bf 72.04 & 90.70 & 71.52 & 90.34 \\
        \midrule
        \bf \ours & 71.88 & 90.50 & \bf 72.65 & \bf 91.08 \\
        \bottomrule
    \end{tabular}
    }
    \vspace{-2mm}
\end{wraptable} 

Note that SSKD~\cite{xu2020knowledge} achieves higher performance because of (1) a better teacher model, and (2) data augmentation for structured knowledge distillation. We further apply \ours~to SSKD as a plug-in by weighting the logit distillation objective and keeping the structured knowledge distillation terms unchanged. Table~\ref{tab:cifar100-sskd} shows that our \ours~can consistently improve SSKD by 0.5\textasciitilde 1.0\% for different architectures. These results indicate that our \ours~is a good complementary to distillation methods.

Table~\ref{tab:imagenet-mobile} further shows the comparison on ImageNet. Following CRD~\cite{tian2019contrastive} and WSLD~\cite{zhou2021rethinking}, we used two teacher-student architectures as the representatives. For the same architecture style, our \ours~improves KD by 1.21\%, and achieves competitive performance compared to WSLD. For the different architecture style, the improvement of WSLD over KD drops from 1.37\% to 1.03\%. As a comparison, our \ours~improves KD by 2.16\%, and outperforms WSLD by 1.13\%. This improvement on the large-scale dataset further demonstrates the effectiveness of our \ours~to bridge the transfer gap when the student and teacher model have different architecture styles, which is more practical in real-world applications.

\begin{table*}
    \caption{Ablation study of technical designs for weight estimation on CIFAR-100. ``CLS head'' denotes the usage of an extra classification head. ``logits norm.'' denotes that the logits are normalized before calculating the propensity.}
    \label{tab:cifar100-ab-component}
    \centering
    \scalebox{0.91}{
    \begin{tabular}{c cc ccc ccc}
    \toprule
    & & & \multicolumn{3}{c}{Same architecture style} & \multicolumn{3}{c}{Different architecture style} \\ 
    \cmidrule(lr){4-6} \cmidrule(lr){7-9} 
     & & & {\scriptsize WRN-40-2} & {\scriptsize resnet110} & {\scriptsize resnet32x4} & {\scriptsize resnet32x4} & {\scriptsize resnet32x4} & {\scriptsize WRN-40-2} \\
     & & & {$\downarrow$} & {$\downarrow$} & {$\downarrow$} & {\scriptsize $\downarrow$} & {$\downarrow$} & {$\downarrow$} \\
     & CLS head & logits norm. & {\scriptsize WRN-40-1} & {\scriptsize resnet32} & {\scriptsize resnet8x4} & {\scriptsize ShuffleNetV1} & {\scriptsize ShuffleNetV2} & {\scriptsize ShuffleNetV1} \\
    \midrule
     & & & \multicolumn{3}{c}{\small training diverges} & 52.81 & 57.99 & 53.31 \\
     & & \cmark & 74.01 & 73.41 & 75.89 & 75.49 & 76.48 & 76.34 \\
     & \cmark & & 74.42 & 73.48 & 75.97 & 75.80 & 76.45 & 75.96 \\
    \midrule
    \bf \ours & \cmark & \cmark & \bf 74.64 & \bf 73.91 & \bf 76.03 & \bf 76.03 & \bf 76.61 & \bf 76.61 \\
    \bottomrule 
    \end{tabular}
    }
\end{table*}

\textbf{Ablation study: technical designs}. As introduced in Section~\ref{sec:method-imp}, we used an extra classifier head to produce CLS-trained output, and normalized the logits to reduce the variance for propensity estimation. Note that WSLD~\cite{zhou2021rethinking} uses the teacher model to estimate the sample weight, and the teacher model is also trained with the cross-entropy loss. Therefore, we considered an alternative that which replaces the classification head with the teacher model to produce the classification-aware output. To evaluate the contribution of logits normalization, we considered an alternative that the logits are not normalized by the standard deviation.
Results in Table~\ref{tab:cifar100-ab-component} verify the contribution of each design. Without the classification head and logit normalization, the training is hard to converge or the performance is much worse. As a comparison, either the classification head or logit normalization helps with stable training. Besides, a combination of both further improves the performance and achieves the best results. The crash of training is due to the high variance of sample weights. Since the teacher model is well pre-trained and has more parameters, it has a larger capacity than the student model. Differently, an extra head with a shared backbone guarantees a similar capacity. Also, the normalization will avoid an extremely large or small CE loss, which further reduces the variance.

\textbf{Teacher trained with label smoothing.} Recent works~\cite{muller2019does,shen2020label} observed that KD performs poorly with label smoothing. Similar to KD, the performance of IPWD drops when the teacher model is trained with label smoothing, but still outperforms KD. However, we found that the improvement of IPWD compared to KD also decreases with label smoothing. For example, on CIFAR-100, given ResNet50 as teacher and MobileNetV2 as student, IPWD outperforms KD by 1.12\% (69.67\% vs. 68.55\%) without label smoothing, but the improvements drops to 0.56\% (66.79\% vs. 66.23\%) with label smoothing. Given resnet32x4 as teacher and ShuffleNetV1 as student, IPWD outperforms KD by 1.52\% (75.79\% vs. 74.27\%) without label smoothing, but the improvement drops to 0.53\% (73.27\% vs. 72.74\%) with label smoothing. We observed that teacher trained with label smoothing produces more balanced predictions compared to teacher trained without label smoothing. Therefore, the results are consistent with our hypothesis that IPWD helps to bridge the transfer gap especially when the context information of teacher is imbalanced.

\subsection{One-stage Self-Distillation}

\textbf{Baseline methods and metrics}. For one-stage self-distillation, we apply our method to the state-of-the-art PS-KD~\cite{kim2021self} method as a plug-in, and consider label smoothing (LS) method and two self-distillation methods, CS-KD~\cite{yun2020regularizing} and TF-KD~\cite{yuan2020revisiting}, as baselines. PS-KD proposed a one-stage framework that progressively distills the knowledge of a model itself to soften the one-hot supervisions as regularization. The knowledge is transferred using a conventional distillation loss. As for metrics, besides top-1 and top-5 accuracy, we follow Kim \etal~\cite{kim2021self} to report expected calibration error (ECE, \%) and the area under the risk-coverage curve (AURC, $\times 10^3$). A low ECE indicates well-calibrated predictions, and a low AURC represents the well-separation of correct and incorrect predictions.

\textbf{Implementation}. 
We follow all the training details of PS-KD for a fair comparison. Specifically, the architectures we considered are ResNet-18~\cite{he2016identity}, ResNet-101~\cite{he2016deep}, ResNeXt-29~\cite{xie2017aggregated} (cardinality=8, width=64), and DenseNet-121~\cite{huang2017densely} (growth rate=32). During training, PS-KD gradually determine how much the student learns from the teacher's knowledge. The formulation is:
\begin{equation}
    \Loss_{ps\text{-}kd}=(1-\alpha_t)\cdot\Loss_{cls}+\alpha_t\cdot\Loss_{dist},\label{eq:pskd}
\end{equation}
where the trade-off parameter $\alpha_t\!=\!\alpha_T\times t/T$, $T$ is the number of total epochs (\eg, 300), $t$ is the current epoch, and $\alpha_T$ is a hyperparameter. Compared to Eq.~\eqref{eq:pskd}, our \ours~applied on PS-KD is formulated as 
\begin{equation}
    \Loss_{ps\text{-}kd\text{~+~}ipw}=(1-\alpha_t)\cdot\Loss_{cls}+\alpha_t\cdot\Loss_{ipw\text{-}dist}\label{eq:ipwpskd}
\end{equation}
Since both the student and teacher models are poor at early epochs, the weight estimation is not accurate at early epochs, which may lead to a worse self-teacher. Therefore, we apply \ours~at the last 75 epochs over the total 300 epochs. 

\begin{table}[t]
\begin{center}
    \caption{Results on CIFAR-100 test set for the one-stage self distillation setting over four architectures. Top-1 and Top-5 indicate the accuracy.}
    \label{tab:cifar100-pskd}
    \scalebox{0.84}{
    \begin{tabular}{lcccc}
        \toprule
        \textbf{Method} & \textbf{Top-1} & \textbf{Top-5} &\textbf{ECE} & \textbf{AURC}\\
        \midrule
        ResNet-18 & {75.82} & {93.10} & {11.84} & {67.65}\\
        + LS & {79.06} & {93.98} & {10.79} & {57.74}\\
        + CS-KD~\cite{yun2020regularizing} & {78.70} & {94.30} & {6.24} & {56.56}\\
        + TF-KD~\cite{yuan2020revisiting} & {77.12} & {93.99} & {11.96} & {61.77}\\
        + PS-KD~\cite{kim2021self} & {79.18} & {94.90} & {1.77} & {52.10}\\
        \hline
        \bf + PS-KD + Ours & {\bf 79.82} & {\bf 95.15} & {\bf 1.39} & {\bf 49.71}\\
        \midrule
        ResNet-101 & {79.25} & {94.72} & {10.02} & {55.45}\\
        + LS & {80.16} & {94.93} & {3.43} & {95.76}\\
        + CS-KD~\cite{yun2020regularizing} & {79.24} & {94.38} & {12.18} & {64.44}\\
        + TF-KD~\cite{yuan2020revisiting} & {79.90} & {94.90} & {6.14} & {58.80}\\
        + PS-KD~\cite{kim2021self} & {80.57} & {95.70} & {6.92} & {49.01}\\
        \hline
        \bf + PS-KD + Ours & {\bf 81.39} & {\bf 95.91} & {\bf 3.19} & {\bf 43.82}\\
        \bottomrule
    \end{tabular}}
    \scalebox{0.84}{
    \begin{tabular}{lcccc}
        \toprule
        \textbf{Method} & \textbf{Top-1} & \textbf{Top-5} &\textbf{ECE} & \textbf{AURC}\\
        \midrule
        DenseNet-121 & {79.95} & {95.01} & {7.34} & {52.21}\\
        + LS & {80.20} & {94.54} & {\bf 0.92} & {91.06}\\
        + CS-KD~\cite{yun2020regularizing} & {79.53} & {93.79} & {13.80} & {73.37}\\
        + TF-KD~\cite{yuan2020revisiting} & {80.12} & {94.90} & {7.33} & {69.23}\\
        + PS-KD~\cite{kim2021self} & {81.27} & {\bf 96.10} & {3.71} & {45.55}\\
        \hline
        \bf + PS-KD + Ours & {\bf 81.60} & {96.04} & {3.48} & {\bf 45.33}\\ 
        \midrule
        ResNeXt-29 & {81.35} & {95.53} & {\bf 4.17} & {44.27}\\
        + LS & {82.40} & {95.77} & {22.14} & {41.92}\\
        + CS-KD~\cite{yun2020regularizing} & {81.74} & {95.63} & {5.95} & {42.11}\\
        + TF-KD~\cite{yuan2020revisiting}& {82.67} & {96.13} & {6.73} & {40.34}\\
        + PS-KD~\cite{kim2021self} & {82.72}  & {96.40} & {9.15} & {39.78}\\
        \hline
        \bf + PS-KD + Ours & {\bf 83.30} & {\bf 96.60} & {4.93} & {\bf 37.49}\\
        \bottomrule
    \end{tabular}}
\end{center}
\end{table}
\textbf{Comparison with baseline methods}. Table~\ref{tab:cifar100-pskd} shows the results of one-stage self-distillation methods over four architectures. Our \ours~can effectively and constantly improve the top-1 accuracy of PS-KD by 0.33\%$\sim$0.82\% with different architectures. Besides, our \ours~siginifantly lowers the ECE and AURC of PS-KD. These results demonstrate the effectiveness of our \ours.

\begin{wraptable}{r}{8.2cm}
\begin{center}
    \vspace{-6mm}
    \centering
    \caption{Ablation study on the start of applying \ours.}
    \label{tab:cifar100-pskd-epoch}
    \scalebox{0.9}{
    \begin{tabular}{lcccc}
        \toprule
        \textbf{Method} & \textbf{Top-1 Acc} & \textbf{Top-5 Acc} &\textbf{ECE} & \textbf{AURC}\\
        \midrule
        PS-KD~\cite{kim2021self} & {82.72}  & {96.40} & {9.15} & {39.78}\\
        \hline
        + \ours~(early) & {82.86} & {96.35} & {8.56} & {38.16}\\
        + \ours~(late) & {\bf 83.30} & {\bf 96.60} & {\bf 4.93} & {\bf 37.49}\\
        \bottomrule
    \end{tabular}}
\end{center}
\end{wraptable}
\textbf{Ablation study: \ours~stage}. We conduct an ablation study to analyze whether \ours~should be started from an early stage (\eg, the beginning of training) or a late stage (\eg, last 1/4 of the epochs). We take ResNeXt-29 as an example. As shown in Table~\ref{tab:cifar100-pskd-epoch}, applying \ours~from the beginning slightly outperforms PS-KD and under-performs the student modal that applies \ours~only at the late stage by large margins. As the student model is poorly trained at the early stage, the weight estimation is inaccurate and hurts the performance of self-teacher. These results indicate that the quality of estimated weight and distillation performance relies on the student model and self-teacher.

\section{Conclusion}

In this paper, 
we point out that conventional KD methods hold an invalid IID assumption and do not properly address the transfer gap between the context-invariant human domain and the context-equivariant machine domain, especially the imbalance knowledge of the teacher model on the transfer set. We further proposed a simple yet effective method, \oursfull~(\ours), to deal with the imbalanced knowledge caused by transfer gap. 
In the future, we will extend our \ours~to (1) tasks beyond classification, like detection and segmentation, and (2) long-tailed tasks where the training samples in the human domain are also non-IID.

\section*{Acknowledgement}
We thank anonymous ACs and reviewers for their valuable discussion and insightful suggestions.
This research is supported by the National Research Foundation, Singapore under its AI Singapore Programme (AISG Award No: AISG2-RP-2021-022) and Alibaba-NTU Singapore Joint Research Institute (JRI).

{
\bibliographystyle{plain}
\bibliography{egbib}
}

\appendix

\section*{Appendix}

\section{Experimental Settings}

\subsection{Two-stage Distillation}

\textbf{Implementation}. We follow the training details of Tian \etal~\cite{tian2019contrastive} for CIFAR-100 and Zhou \etal~\cite{zhou2021rethinking} for ImageNet. Specifically, for CIFAR-100, we set the mini-batch size as 64 and an initial learning rate as 0.05. We train the model for 240 epochs. The learning rate is decayed by 10 every 30 epochs after 150 epochs. We initialize the learning rate as 0.01 for MobileNetV2, ShuffleNetV1 and ShuffleNetV2, and as 0.05 for other models. The experiments are conducted using one NVIDIA TITAN RTX GPU. For ImageNet, we train the model for 100 epochs. We set the mini-batch size as 256, an initial learning rate as 0.1, and decay it by 10 every 30 epochs. The experiments are conducted using four Tesla V100 GPUs.

\noindent \textbf{Architectures}. We follow~Tian \etal~\cite{tian2019contrastive} for the choice of network architectures. Specifically, ``WRN-d-w'' denotes Wide Residual Network (WRN)~\cite{zagoruyko2016wide} with depth $d$ and width factor $w$. ``resnet-d'' represents cifar-style ResNet~\cite{he2016deep} with 3 groups of basic blocks, each with 16, 32, and 64 channels respectively. For example, resnet8x4 and resnet32x4 represent a 4 times wider network, \ie, with 64, 128, and 256 channels respectively.
``ResNet-d'' represents the ImageNet-style ResNet with Bottleneck blocks and more channels. MobileNetV2~\cite{sandler2018mobilenetv2} has a width multiplier of 0.5.
``vgg'' denotes VGGNet~\cite{simonyan2014very} that is adapted from its original ImageNet counterpart.
``ShuffleNetV1''~\cite{zhang2018shufflenet}, ``ShuffleNetV2''~\cite{ma2018shufflenet} are adapted with input size as 32x32.

\subsection{One-stage Self-Distillation}

\noindent \textbf{Implementation}. We follow all the training details of PS-KD~\cite{kim2021self}. The standard data argumentation schemes are 32x32 random crop after padding with 4 pixels and random horizontal flip. The networks are trained for 300 epochs using SGD with a momentum of 0.9. The learning rate is decayed by 10 at 150 and 225 epochs. We set the mini-batch size as 128, an initial learning rate as 0.1, and a weight decay as 0.0005. The experiments are conducted using four Tesla V100 GPUs.

\noindent \textbf{Metrics}. Besides top-1 and top-5 accuracies, we follow Kim \etal~\cite{kim2021self} and report Expected calibration error (ECE) and Area under risk-coverage curve (AURC) for evaluation. Expected calibration error (ECE)~\cite{naeini2015obtaining} is used to evaluate the confidence calibration performance of models, \ie, the expected gap between accuracy and confidence. Specifically, the samples are partitioned by confidence into $M$ bins $B_1,\cdots,B_M$. The $m$-th bin $B_m$ contains samples with confidence within $[\frac{m-1}{M}, \frac{m}{M}]$. For $N$ samples, ECE is formulated as:
\begin{equation*}
    \text{BCE}=\frac{1}{N}\sum_{m=1}^{M}|B_m|\times|\text{Acc}(B_m)-\text{Conf}(B_m)|,
\end{equation*}
where $\text{Acc}(B_m)$ denotes the accuracy of samples in $B_m$, and $\text{Conf}(B_m)$ denotes the average confidence of samples in $B_m$. A lower BCE indicates a well-calibrated model.

Area under risk-coverage curve (AURC)~\cite{geifman2018bias} measures how well predictions are ordered by confidence values. Specifically, we can determine a threshold for classification, where only samples with confidence higher than the threshold are accepted. After that, we can obtain the proportion of covered samples to the whole dataset, \ie, coverage, and define the risk as the error rate computed by using the covered samples. AURC is defined as the area under the risk-coverage curve. A lower AURC indicates that the correct and incorrect predictions are well-separable by confidence values.

\subsection{License of Assets}
We reimplemented WSLD\footnote{\url{https://github.com/bellymonster/Weighted-Soft-Label-Distillation}} and SSKD\footnote{\url{https://github.com/xuguodong03/SSKD}} based on their open-resourced codes. Both WSLD and SSKD did not mention the license in their open-resourced codes.

\section{Additional Results}

\subsection{Architecture styles}
Due to page limitation, we did not include all the results of different architecture styles on CIFAR-100 in the main paper. We provide additional results in Table~\ref{tab:cifar100-additional}. Results show that our \ours~achieves competitive performances and outperforms KD by large margins.

\begin{table*}
    \caption{Top-1 accuracy of student networks on CIFAR-100 test set for the two-stage distillation setting. $^*$ denotes our reproduced results using the same teacher model.}
    \label{tab:cifar100-additional}
    \vspace{-2mm}
    \centering
    \scalebox{0.96}{
    \begin{tabular}{l cc cc}
    \toprule
    & \multicolumn{2}{c}{Same architecture style} & \multicolumn{2}{c}{Different architecture style} \\ 
    \cmidrule(lr){2-3} \cmidrule(lr){4-5} 
    Teacher & {\scriptsize WRN-40-2} & {\scriptsize resnet110} & {\scriptsize resnet32x4} & {\scriptsize vgg13}\\
    Student & {\scriptsize WRN-16-2} & {\scriptsize resnet20} & {\scriptsize ShuffleNetV2} & {\scriptsize MobileNetV2}\\
    \midrule
    Teacher & 75.61 & 74.31 & 79.42 & 74.64 \\
    Student & 73.26 & 69.06 & 71.82 & 64.60 \\ 
    \midrule
    FitNet~\cite{romero2014fitnets} & 73.58 & 68.99 & 73.54 & 64.14\\
    AT~\cite{zagoruyko2016paying} & 74.08 & 70.22 & 72.73 & 59.40 \\
    SP~\cite{tung2019similarity} & 73.83 & 70.04 & 74.56 & 66.30\\
    CC~\cite{peng2019correlation} & 73.56 & 69.48 & 71.29 & 64.86\\
    VID~\cite{ahn2019variational} & 74.11 & 70.16 & 73.40 & 65.56\\
    RKD~\cite{park2019relational} & 73.35 & 69.25 & 73.21 & 64.52\\
    PKT~\cite{passalis2018learning} & 74.54 & 70.25 & 74.69 & 67.13\\
    AB~\cite{heo2019knowledge} & 72.50 & 69.53 & 74.31 & 66.06\\
    FT~\cite{kim2018paraphrasing} & 73.25 & 70.22 & 72.50 & 61.78\\
    NST~\cite{huang2017like} & 73.68 & 69.53 & 74.68 & 58.16\\
    KD~\cite{hinton2015distilling} & 74.92 & 70.67 & 74.45 & 67.37\\
    CRD~\cite{tian2019contrastive} & 75.48 & \bf 71.46 & 75.65 & 69.73\\
    WSLD$^*$~\cite{zhou2021rethinking} & 75.63 & 71.20 & 75.55 & 68.50 \\
    \midrule
    \bf \ours & \bf 75.83 & 71.22 & \bf 76.61 & \bf 69.81 \\
    \bottomrule 
    \end{tabular}
    }
\end{table*}

\subsection{Feature-based Methods}

\begin{table}
    \caption{{ReviewKD with our IPWD reweighting strategy on feature level.}}
    \label{tab:cifar100-feature}
    \centering
    \vspace{2mm}
    \scalebox{0.99}{
    \begin{tabular}{l ccccc}
    \toprule
    Teacher & {\scriptsize WRN-40-2} & {\scriptsize resnet56} & {\scriptsize resnet110} & {\scriptsize resnet32x4} & {\scriptsize WRN-40-2}\\
    Student & {\scriptsize WRN-16-2} & {\scriptsize resnet20} & {\scriptsize resnet32} & {\scriptsize ShuffleV2} & {\scriptsize ShuffleV1} \\
    \midrule
    ReviewKD        & 76.12 & \bf 71.89 & \bf 73.89 & \bf 77.78 & \bf 77.14 \\
    ReviewKD + IPWD & \bf 76.25 & 71.51 & 73.79 & 77.74 & 77.06 \\
    \bottomrule 
    \end{tabular}
    }
\end{table}

Note that our proposed IPWD is a logit-based distillation method. An interesting question is whether the reweighting strategy can work with feature-based distillation methods. We select ReviewKD as an example, which is a recent representative feature-based distillation method. As shown in Figure~\ref{tab:cifar100-feature}, the gap between ReviewKD+IPWD and ReviewKD is very marginal, which indicates that IPWD cannot promote feature-based distillation. The possible reasons are two-fold. First, the logit knowledge of label $y$ is long-tailed but the representation knowledge of sample $x$ may be relatively balanced. Second, as pointed out by Kang \etal~\cite{kang2019decoupling}, ``data imbalance might not be an issue in learning high-quality representations'' for long-tailed classification, which implies that the reweighting strategy is not compatible at feature level.

\subsection{{Long-tailed Methods for KD}}

{Note that Figure~\ref{fig:teaser} shows that there seems long-tailed property of teacher predictions. An interesting question is whether it could be fixed by techniques for long-tailed classification. We take LA~\cite{menon2020long} as the recent representative technique for long-tailed classification. LA proposed a logit adjusted softmax cross-entropy loss by applying a class prior to each logit. LA does not require extra modules (compared to TDE~\cite{tang2020long}), post-hoc logit adjustment (compared to LADE~\cite{hong2021disentangling}), or ensemble of multiple models (compared to RIDE~\cite{wang2020long}).}

{Following LA, we applied the class prior to the student output when calculating the KL divergence distillation loss. We found that KD+LA underperforms KD by averagely 0.5\% on CIFAR-100. The possible reason is that the introduced prior indirectly breaks the teacher's knowledge for each training sample, which hurts the effectiveness of distillation. These results indicate that logit-adjust-based long-tailed techniques are not applicable to the issue of KD.}

\subsection{Ablation Studies}

We have conducted ablation studies on the components of our proposed \ours. In the appendix, we further provide ablation studies on the technical designs.

We take two-stage distillation on CIFAR-100 as an example. Recall that the sample weights are estimated based on the two types of student's outputs, KD-trained output $y^{kd}$ from the student's original head and CLS-trained output $y^{cls}$ from the student's extra head. We further conduct ablations on the selection of two outputs.

For CE-aware output $y^{cls}$, an straightforward alternative is using an extra model that has the same architecture as student with totally different parameters. This extra model is trained using the cross-entropy classification loss. In other words, the difference is whether the visual backbone is shared for the two outputs. We denote this alternative as \ours$^*$. Table~\ref{tab:cifar100-ab-cls} shows the comparison. Overall, \ours$^*$ achieves competitive results compared to \ours. However, training an extra model leads to more memory and time cost. Therefore, our design that takes an extra head is both effective and efficient.

For KD-trained output $y^{kd}$, an straightforward alternative is using an extra distillation head like the classification head. Different from the original head of the student, the distillation head is trained only using the distillation loss, which simply mimics the teacher's output without ground-truth annotations. We denote this alternative as \ours$^\dag$. Table~~\ref{tab:cifar100-ab-kd} shows the comparison. Overall, \ours$^\dag$ slightly underperforms \ours~with different four architectures. The reason is that the extra distillation head only learns from the teacher model, which is sensitive to the teacher's performance and weights hard samples more. These two ablation studies further verify the effectiveness of our techinical designs.

\begin{table}
    \caption{Ablation study of CE-aware output on CIFAR-100. 
    $^*$ denotes that the CE-aware output is obtained from an extra student model.
    }
    \label{tab:cifar100-ab-cls}
    \centering
    \vspace{2mm}
    \scalebox{0.99}{
    \begin{tabular}{l cc cc}
    \toprule
    Teacher & {\scriptsize resnet110} & {\scriptsize resnet32x4} & {\scriptsize resnet32x4} & {\scriptsize resnet32x4} \\
    Student & {\scriptsize resnet32} & {\scriptsize resnet8x4} & {\scriptsize ShuffleV1} & {\scriptsize ShuffleV2} \\
    \midrule
    Teacher & 74.31 & 79.42 & 79.42 & 79.42 \\
    Student & 71.14 & 72.50 & 70.50 & 71.82 \\
    \midrule
    \ours$^*$ & 73.64 & 75.88 & 75.98 & 76.83 \\
    \bf \ours & 73.91 & 76.03 & 76.03 & 76.61 \\
    \bottomrule 
    \end{tabular}
    }
\end{table}

\begin{table}
    \caption{Ablation study of KD-aware output on CIFAR-100. 
    $^\dag$ denotes that the KD-aware output is obtained from another extra head trained only with the original distillation loss.
    }
    \label{tab:cifar100-ab-kd}
    \centering
    \vspace{2mm}
    \scalebox{0.99}{
    \begin{tabular}{l cc cc}
    \toprule
    Teacher & {\scriptsize resnet110} & {\scriptsize resnet32x4} & {\scriptsize resnet32x4} & {\scriptsize resnet32x4} \\
    Student & {\scriptsize resnet32} & {\scriptsize resnet8x4} & {\scriptsize ShuffleV1} & {\scriptsize ShuffleV2} \\
    \midrule
    Teacher & 74.31 & 79.42 & 79.42 & 79.42 \\
    Student & 71.14 & 72.50 & 70.50 & 71.82 \\
    \midrule
    \ours$^\dag$ & 73.56 & 75.88 & 75.93 & 76.44 \\
    \bf \ours & 73.91 & 76.03 & 76.03 & 76.61 \\
    \bottomrule 
    \end{tabular}
    }
\end{table}

\end{document}